# Accurate Segmentation of Dermoscopic Images based on Local Binary Pattern Clustering

Pedro M. M. Pereira[1,3], Rui Fonseca-Pinto[1,2], Rui Pedro Paiva[3], Luis M. N. Tavora[2],
Pedro A. A. Assuncao[1,2], Sergio M. M. de Faria[1,2]

[1] Instituto de Telecomunicações, Portugal
[2] ESTG, Polytechnic Institute of Leiria, Portugal
[3] DEI - FCTUC, University of Coimbra, Portugal
pedrommpereira@co.it.pt

*Abstract* - Segmentation is a key stage in dermoscopic image processing, where the accuracy of the border line that defines skin lesions is of utmost importance for subsequent algorithms (e.g., classification) and computer-aided early diagnosis of serious medical conditions. This paper proposes a novel segmentation method based on Local Binary Patterns (LBP), where LBP and K-Means clustering are combined to achieve a detailed delineation in dermoscopic images. In comparison with usual dermatologist-like segmentation (i.e., the available ground-truth), the proposed method is capable of finding more realistic borders of skin lesions, i.e., with much more detail. The results also exhibit reduced variability amongst different performance measures and they are consistent across different images. The proposed method can be applied for cell-based like segmentation adapted to the lesion border growing specificities. Hence, the method is suitable to follow the growth dynamics associated with the lesion border geometry in skin melanocytic images.

*Keywords - Segmentation, Lesion Detection, Medical Imaging, Dermoscopy*

## I. Introduction

Almost any cell in the body can develop cancer, and in the case of skin cells, melanoma is the deadliest occurring type [1]. By using digital technology (*e.g.* dermoscopy) and image processing techniques, it has been possible to early detect suspicious skin lesions and, when detected at an early stage, melanoma presents high cure rates [2].

Early detection of specific features of each type of lesion by image processing algorithms enables the use of Machine Learning (ML) techniques with great success in the automatic classification of dermoscopic images [3]. These specific features used to feed ML algorithms are usually divided in two groups according to its origin (*i.e.*, lesion or skin). Accordingly, amongst all image processing steps commonly used in dermoscopic images, the identification of the region of interest (ROI) is of central importance in the classification framework [4]. In addition to the ROI delineation, this segmentation procedure is also used not only to extract other information regarding the lesion itself, but also about the dynamics of its growing process [5].

The manual (round-like) segmentation obtained by dermatologists, and used as ground-truth in the majority of image datasets, is mostly performed to identify surgical borders for excision, lacking an objective rule or metrics. Moreover, variations in lightning conditions can influence contrast and blur, thus precise identification of skin lesion boundaries poses a problem to manual segmentation [6]. Even when clinicians are guided to perform a cell-like based delineation of the lesion, this procedure has proven to suffer from high inter- and intra-observer variability [6-8]. As a consequence, the resulting ground-truth, cell-based like, hand segmentation lacks definiteness.

In the literature, a broad range of segmentation algorithms mostly covering the above mentioned round-like segmentation have been proposed, ranging from smoothing and thresholding, to color space conversions, and exploiting specific aspects of skin dermoscopic images as reported in [9]. In fact, this wide range of methodologies is related with the datasets diversity regarding physical acquisition conditions (e.g., light, angle of view), anatomical and local artifacts (e.g., hairs, skin curvatures) and equipment properties (e.g., lens, light, image resolution) [10, 11].

In some recent proposed segmentation methodologies [12-14] the results achieve accuracy levels above 90%. Additionally, in other works, preprocessing techniques have been proposed (illumination correction, contrast enhancement and hair removal) to improve similarity indexes. In particular RGB color space conversion to *CIE L\*a\*b\** was proposed in [15] to ease hair removal prior to segmentation and in [16] an averaging filter was applied to the luminance channel. These recent methodologies for segmentation reported overlapping scores of 90% and 83%, respectively, showing a high degree of similarity between ground-truth and automatic segmentations. Other recent threshold based approaches have also emerged, such as iterative thresholding [17], threshold fusion [18, 19] or hybrid thresholding [20]. In addition, other methods have also been proposed, combining different categories, such as clustering [21, 22], soft computing (neural networks [23, 24] and evolution strategy [25]), supervised learning [26], active contours [27] and gradient [28].

Although cell-like based delineation is quite useful for assessment of lesion growing dynamics, it is difficult to obtain a ground-truth reference for each image. Note that this type of segmentation is absent in datasets and the manual delineation can be influenced by external factors.

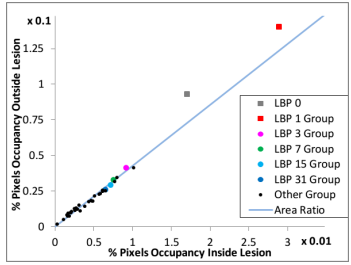

Figure 1. Invariant LBPs Presence Ratio for image IMD021 [35].

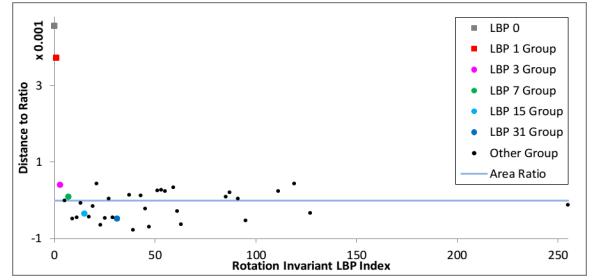

Figure 2. Invariant LBP significance for image IMD021 [35].

This work presents a contribution to overcome the previously described shortcomings, by proposing an algorithm for cell-based like segmentation based on Local Binary Patterns (LBP) and clustering, which is independent of human factors.

This paper is organized as follows: Section II resumes the algorithm background and in Section III the proposed method is presented. In Section IV the experimental results are presented and the work is concluded with a discussion and conclusions part in Section V.

## II. LOCAL BINARY PATTERNS

In general, regions of normal skin in dermoscopic images present flatter texture when compared to regions within the lesion. This characteristic can be exploited in order to identify those different spots by using LBPs [29].

### A. LBP and Rotation Invariants

The LBP operator is a 2D texture descriptor that assesses local variations on the image, and codes them in terms of a spatial pattern with an associated grayscale scheme. The underlying idea behind LBP operators is that texture has locally two complementary aspects: a spatial pattern and a corresponding strength. In fact, LBPs can be seen as an image operator, whose output is an array of integer labels describing small-scale variations (high frequency content) in the image. These labels, or their statistics, can then be used for further image analysis.

One simple variant of classical LBP as presented in [29] is defined in (1) for each pixel image pixel.

$$\text{LBP} = \sum_{p=0}^{7} s(I_p - I_c) \cdot 2^p, \quad s(x) = \begin{cases} 1 \text{ if } x > 0 \\ 0 \text{ if } x \leq 0 \end{cases} \quad (1)$$

In the process of obtaining the LBP, each pixel $I_C$ is compared to its neighbors $I_P$ (i.e., 8 pixels $I_P$ surrounding the central pixel $I_C$ in a 3x3 arrangement) producing a binary number with 8 digits. According to (1), this number is then converted to a decimal base (which, in the case of 8-bit images, conducts to LBPs ranges from 0 to 255). The histogram of the obtained LBPs can then be used to generate image features.

The direct application of (1) leads to a high number of possible combinations of binary patterns, hence other derivations of this methodology have been proposed in the literature depending on the types of images and processing objectives. One of the possible formulations is to select only the patterns that are invariant to rotations, since this reduces the number of patterns and also because these invariants are associated to some geometric primitives within an image (corners, edges, flat regions, dark and bright spots) [30]. Apart from the 00000000 and the 11111111 LBP sequences, almost all binary patterns can be visually rotated by applying a binary shift. As an example, 01110000 and 00111000 are rotation invariants. This creates subsets of patterns that can be accounted for by one representative of each invariant class, as in [30].

This type of invariant LPBs can still be refined to achieve a set of rotation invariant patterns, containing a specific number of transitions in the binary sequence (between 0s and 1s and vice-versa). For each group of the same invariant LBPs, if $n$ is the smallest decimal number within the group then the representative of the class is labeled as LBP$n$. In particular, when the number of transitions is at most two, this set is called uniform pattern of LBP (uLBP) and the total number of uLBPs are 58, considering 8 neighbors.

### B. LBP Invariants in Dermoscopy

The number of binary transitions within an invariant LBP is related to its ability to discriminate between different texture patterns. In fact, the larger the number of transitions the more likely is the change to a different pattern upon rotation in digital domain [30]. Accordingly, a plausible hypothesis is that a reduced number of transitions allow to capture local texture variations and also to identify the locations where such variations are mostly regular (i.e. the segmentation borderline). This hypothesis was tested by means of a set of experiments by assigning LBPs either to skin or to lesion, according to the texture properties of the corresponding image regions.

Firstly, for each image, the ratio between the lesion area and the whole image was obtained, by using the ground-truth segmentation and the number of pixel in each region. This defines the slope of a reference line in plane (x,y), where x and y represent the percentage of each invariant LBP inside and outside the lesion considering the whole image. For one dermoscopic image (Fig.1), the presence of each type of LBP$n$ ($n$=0, 1, 3, 7, 15, 31 and others in black) below or above the reference line indicates their dominance in either lesion or normal skin, respectively. LBPs that are positioned above the reference line are dominant outside the lesion area, while LBPs positioned below are dominant inside the lesion area. Additionally, the greater the distance from the origin the greater the presence of a given LBP in the image.

By analyzing Fig. 1, it becomes clear that LBP1 and LBP0 are the densest type of LBPs, both belonging to the normal skin region. For a better understanding, data in Fig. 1 was rotated (according to the reference slope) to align the reference line with the x-axis and then the distance of each LBP group to the line in Fig. 1 appears as presented in Fig. 2. In this figure, the x-axis defines LBP group pattern number as its index.

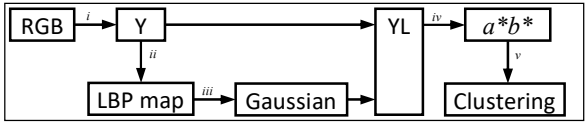

Figure 3. Processing work flow: a given RGB image is converted into greyscale (Y) whereat a LBP map is extracted and passed through a Gaussian filter (L); then Y and L are mapped together and converted into the $a^*b^*$ coordinate color space; finally a clustering algorithm processes the space into two clusters.

This representation reinforces the conclusion that rotation invariants LBP0 (formed by LBP '0') and LBP1 groups (formed by LBPs of '1', '2', '4', '8', '16', '32', '64' and '128', also known as LBPs of power of 2 or the LBP1 group) are more present in the normal skin region where the aforementioned property regarding flat textures exists. This result can be mostly reproduced for the other dermoscopic images from the dataset, and can be used for several purposes. In particular, these invariant LBPs can be used to support segmentation methodologies, which is the main subject of this work.

## III. PROPOSED METHOD

The segmentation method herein described exploits the fact that lesion and healthy skin tend to exhibit different texture pattern variations. In particular, and given the discussion carried out in Section II, a lower density of LBPs of the groups '0' and '1' is used as an indicator to identify lesion areas. That information, together with the image luminance, which, on its own is known to be a reliable indicator of the lesion area [31], is then fed to a K-Means clustering algorithm.

The overall method proposed for lesion segmentation based on LBP clustering is depicted in Fig. 3, where the main processing steps are: *i)* conversion of the RGB image to greyscale, with luminance Y determined from (2); *ii)* determination of LBPs groups '0' and '1' and the corresponding L binary map; *iii)* gaussian smoothing the L map and fitting in the [0-255] range; *iv)* space conversion from YL to $a^*b^*$; *v)* and group data onto 2 clusters using K-Means.

Then some morphological operations are applied to ensure that no small artifacts exist within the lesion perimeter or, in contrast, in the surrounding skin (which incorrectly masks the area with holes).

### A. Luminace and LBP data

For a given dermoscopic image in RGB format, the corresponding luminance ($Y$) is obtained following the ITU recommendation [32], expressed in (2), and this information is then used to determine the associated LBP as given by (1). Then a binary map L is built based on LBPs from the groups '0' and '1' as follows: pixels with LPBs from these groups are mapped as L=0, while for those with no such LBPs are mapped as L=1. Following the previous arguments, L=0 is expected to be dominant in healthy skin regions while L=1 should dominate in the lesion region.

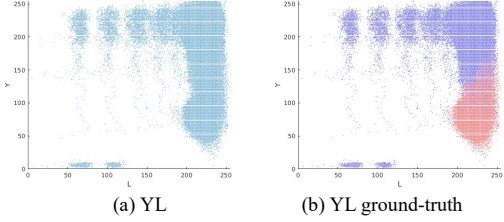

Figure 4. Luminance (Y) and LBP indicator (L) for the dermoscopic image IMD078 [35] (L was previously smoothed using a Gaussian filter and converted to a [0-255] scale). Points were associated to either lesion (red) or healthy skin (blue) based on the provided ground-truth segmentation.

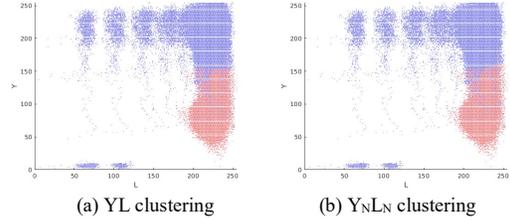

Figure 5. Clustering of YL data (image IMD078 [35]) in its original format (a) and after normalisation (b). Points color represent their cluster association.

$$Y = 0.2989 \times R + 0.5870 \times G + 0.1140 \times B \quad (2)$$

A representation of both Y and L, as shown in Fig. 4, makes it clear about the relevance of such a joint analysis: pixels associated with the lesion are mapped near the lower right corner (*i.e.* high L and low Y) while healthy skin scatters over the top of the diagram (high/low L and high Y). Since the kind of mapping shown in Fig. 4 is characteristic of dermoscopic images, the segmentation challenge can be seen as how to accurately group the pixels associated to the lesion.

### B. Clustering

To provide supporting evidence for the method proposed to convert the YL space into the $a^*b^*$, the following study was carried out. The image data as represented in the form of YL maps (see Fig. 4) was fed into the clustering algorithm to assign pixels either to lesion or healthy skin. The results obtained with the K-means, presented in Fig. 5a, show a limited accuracy when compared with the ground truth (Fig. 4b).

Since the clustering performance can be severely affected by data sets with different scales and/or variance [33], a further normalization of the data was implemented for the YL representation, as expressed in (3) and (4). This leads to data sets ($Y_N$, $L_N$), both with zero mean and unit variance. Clustering these normalized data sets leads to what is shown in Fig. 5b, which seems in much better agreement with ground truth data.

$$Y_N = (Y - Y_\mu)/Y_\sigma \quad (3)$$

$$L_N = (L - L_\mu)/L_\sigma \quad (4)$$

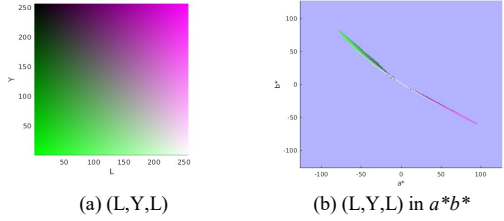

Figure 6. Visualization of Y representing pink and L representing green (a) and its representation in the $a*b*$ space of the *CIE L*a*b** color space (b).

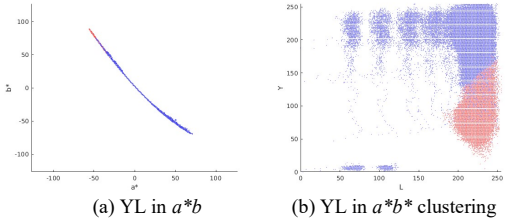

Figure 7. Data visualization for image IMD078 [35] in the $a*b*$ space (a) and its resulting K-Means clustering (b). Points color represent their cluster association.

Moreover, a further representation of the YL data, prior to clustering, was also considered to provide a better discrimination. This representation is inspired on the *CIE L*a*b** color space [34], where colors like green and pink are kept well further apart. If each point in the YL space is assigned a *RGB* color given by (*L*,*Y*,*L*), the color distribution shown in Fig. 6a is obtained. Therefore, applying the *RGB* to *CIE L*a*b** transformation leads to the $a*b*$ representation shown in Fig. 6b. As expected, such operation keeps the top left and the lower right corners of the YL space well separated. Since this space conversion has implicit normalization, it also reduces the data spatial spread. Fig. 7a presents the $a*b*$ representation of YL dermoscopic data, with Fig. 7b showing the promising results of a K-means based clustering. In the proposed method, this space conversion is implemented before clustering, as shown in Fig. 3.

## IV. EXPERIMENTAL RESULTS AND DISCUSSION

The proposed methodology was tested on a set of 200 dermoscopy images (80 common nevi, 80 atypical nevi, and 40 malignant melanomas) which comprise the PH[2] dataset [35] of 765x572 pixels. A MSI GT Series GT683DXR-603US Laptop (Intel Core i7 2nd Generation 2670QM 2.20 GHz, 64-Bit CPU, 12 GB of Memory RAM) was used to execute and calculate the mean execution time of the proposed method, which is of 1.16 seconds on CPU.

The dermoscopic images (8-bit RGB color images) were obtained under the same conditions through Tuebinger Mole Analyzer system with a 20x magnification factor. The database includes the ground-truth segmentation masks that were used to assess the performance of the proposed segmentation algorithm.

Three measures were applied to compare the segmentation provided by the proposed algorithm with the one obtained by the ground-truth masks.

TABLE I. RESULTS FOR PH[2] DATASET

| | Complete Dataset | | | Filtered Subset | | |
|---|---|---|---|---|---|---|
| | BE | TDR | FPR | BE | TDR | FPR |
| $\mu$ | 37.572 | 67.892 | 18.457 | 14.156 | 88.400 | 5.224 |
| $\sigma$ | 33.163 | 31.985 | 24.057 | 4.731 | 6.152 | 3.719 |
| CV | 0.883 | 0.471 | 1.303 | 0.334 | 0.069 | 0.713 |

The Border Error (BE) metric, displayed in percentage, is defined in (5), measures the non-overlapping segmentation regions between the proposed segmentation method (SM) and the dataset ground-truth segmentation (GT).

$$BE(SM,GT) = \frac{Area(SM \oplus GT)}{Area(GT)} \times 100 \quad (5)$$

To quantify incorrect segmentation (skin versus lesion) the True Detection Rate (TDR) defined in (6) and the False Positive Rate (FPR) defined in (7) were also applied acting as a discriminative factor over the Border Error metric. The TDR measures the ratio of pixels that are correctly classified as lesion and the FPR measures the ratio of pixels that are incorrectly classified as lesion.

$$TDR(SM,GT) = \frac{np(SM \cap GT)}{np(GT)} \quad (6)$$

A concise inspection of the whole database allows us to identify few tricky images; in which any segmentation algorithm will certainly fail (e.g. some lesions are not completely captured by the image). Hence, two datasets were tested. The original PH[2] dataset, and the one obtained by removing the images where the border vanishes or where hairs are present, the so-called filtered dataset.

Statistical descriptors were calculated for each of the above defined metrics (5-7) for the two datasets, in particular the mean ($\mu$), standard deviation ($\sigma$), and coefficient of variation (*CV*). The results can be found in Table I.

The results in Table I confirm the expected superior performance for all the metrics when the filtered subset is used for segmentation. In fact, by comparing the figures between both sides of Table I, it is possible to observe a reduced dispersion of data at the filtered dataset. As an example, FPR *CV* in the whole dataset is bigger than one indicating $\sigma > \mu$. This massive data dispersion is reinforced by the box-plots presented in Fig. 8a and Fig. 8b, highlighting the reduced variability amongst data from the filtered dataset.

The statistics for the filtered dataset, in particular the reduced BE and high TDR, points to a segmentation algorithm having good performance. Comparisons with other segmentation methods cannot be straightforwardly done without taking into consideration that the segmentation method proposed in this work was designed to obtain fine segmentation borders able to follow the irregular pattern of a cell-based like growing. However, most of the available algorithms aim to obtain segmentation masks as close as possible to the ground truth, which in general do not have detailed boundaries as pointed out in Section I. The segmentation matching between the output of an automatic procedure and the one obtained by human experts must be interpreted with caution. The ground-truth segmentation is mostly obtained having predefined assumptions (e.g. the definition of surgical margins) thus targeting different objectives. In [28] a gradient-based metric ($G_\perp$) is presented to assess the accuracy of a segmentation border based on the rationale that the segmentation contour is expected to separate regions with substantially different tonalities. Accordingly, the higher the image gradient at a delineation line, the higher the confidence in the delimitation of the lesion area. The results indicate that, on average, the proposed method attains a $G_{\perp,\text{Prop}}$ of 60.018 on the filtered subset. In comparison to [28], the proposed method achieves a quotient gradient $G_{\perp,\text{Prop}} / G_{\perp,GT}$ of 1.593 for the PH$^2$ dataset and 1.891 on the filtered subset, showing the efficiency of the segmentation method while still providing contour detail.

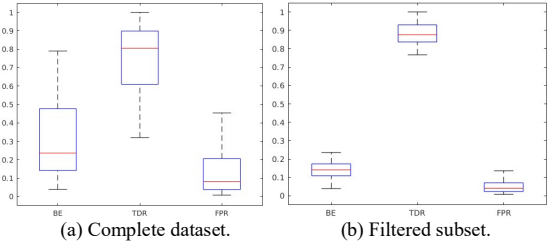

(a) Complete dataset.    (b) Filtered subset.

Figure 8.    Segmentation metrics boxplot for the PH$^2$.

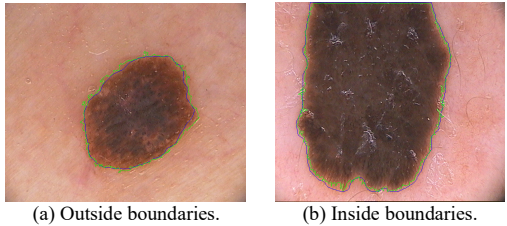

(a) Outside boundaries.    (b) Inside boundaries.

Figure 9.    Example of an accurate segmentation.

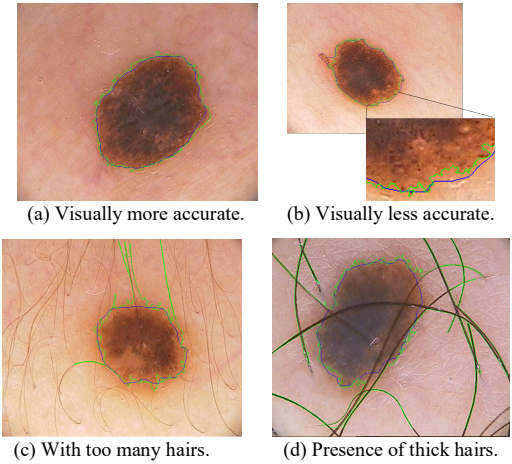

(a) Visually more accurate.    (b) Visually less accurate.

(c) With too many hairs.    (d) Presence of thick hairs.

Figure 10.    Example of an inaccurate segmentation.

Fig. 9a and Fig. 9b are examples of good segmentation results where the algorithm closely matches the ground truth segmentation but provides extra detail to the lesion border thus generating small border errors that translate to increases in FPR and decreases in TDR. Both Fig. 10a and Fig. 10b are examples of bad segmentation. Fig. 10a visually seems segment closer to the lesion core while Fig. 10b actually misses a region of the lesion (upper-left stem). Fig. 10c and Fig. 10d both depict results when hairs are present near the lesion area. In Fig. 10c there are many hairs passing through the lesion and the border segmentation gets elongated along the darker groups of hair. In Fig. 10d the presence of thick hairs makes provides the worst case scenario result.

To investigate the algorithm performance within the type of lesions in the dataset (Atypical Nevus-AN; Common Nevus-CN and Melanoma-M) the previously defined metrics and statistics were calculated and shown in Table II, for the filtered subset.

TABLE II.    EXPANDED RESULTS FOR FILTERED SUBSET

|   | BE | | | TDR | | | FPR | | |
|---|---|---|---|---|---|---|---|---|---|
|   | AN | CN | M | AN | CN | M | AN | CN | M |
| $\mu$ | 13.287 | 15.159 | 13.864 | 89.529 | 87.085 | 88.870 | 4.789 | 5.290 | 7.472 |
| $\sigma$ | 5.189 | 4.334 | 2.970 | 6.398 | 5.897 | 5.186 | 3.380 | 3.968 | 3.884 |
| CV | 0.391 | 0.286 | 0.214 | 0.071 | 0.068 | 0.058 | 0.706 | 0750 | 0.520 |

Globally it is possible to observe different results for each type of dermoscopic image. Additionally, there is a trend associating lower CV values to some degree of malignancy. In fact, CV values for melanoma indexes are the lowest for any type of lesion. This result should be regarded with some caution and confirmed by testing other datasets.

## V.    CONCLUSION

In this paper an unsupervised approach for accurate border detection in dermoscopy images, based on LBP sequences and K-means clustering is presented and validated using a certified dataset and the provided human blind segmentation. The proposed methodology comprises of 3 main phases: LBP image enhancement, $YL$ to $a^*b^*$ space transformation, and clustering.

To assess the performance of the proposed method three metrics were tested (BE, TDR and FPR) jointly with the standard statistics. In a first analysis, the results clearly show the dependence of the algorithm from the dataset quality, thus a filtered subset was used by removing low quality images.

The results regarding the segmentation for this new filtered dataset show reduced variability amongst the defined performance measures, thus indicating the good segmentation consistency across all images. Moreover, the analysis of the filtered dataset by image type indicates a CV decreasing in the Melanoma group. This finding will be probably related to the type of segmentation of the ground truth, and puts forward the need for using different approaches to assess automatic segmentation.

The geometric shape provided by the proposed segmentation algorithm is suitable to identify the real lesion border (instead of the common smooth lines) and can be further applied for a cell-based like segmentation adapted to the lesion border growing. Once this growing process is distinct in different types of lesions (e.g. in particular in the melanoma group) the ability to extract this feature with an algorithm having the aforementioned properties can be used to classify skin lesions based on cell-based like segmentation. This work will be done in the near future.

ACKNOWLEDGMENT

This work was supported by the Fundação para a Ciência e Tecnologia, Portugal, under PhD Grant SFRH/BD/128669/2017 and project PlenoISLA PTDC/EEI-TEL/28325/2017, in the scope of R&D Unit 50008, through national funds and where applicable co-funded by FEDER – PT2020.